\documentclass[conference]{IEEEtran}
\IEEEoverridecommandlockouts
\usepackage{cite}
\usepackage{amsmath,amssymb,amsfonts}
\usepackage{algorithmic}
\usepackage{graphicx}
\usepackage{textcomp}
\usepackage{xcolor}
\usepackage{array}
\usepackage{makecell}
\usepackage{multirow}
\usepackage{subfig}
\usepackage{enumitem}
\def\BibTeX{{\rm B\kern-.05em{\sc i\kern-.025em b}\kern-.08em
    T\kern-.1667em\lower.7ex\hbox{E}\kern-.125emX}}
\begin{document}

\title{Predicting Satisfied User and Machine Ratio for Compressed Images: A Unified Approach}

\author{
    Qi Zhang$^1$, Shanshe Wang$^1$, Xinfeng Zhang$^2$, Siwei Ma$^1$, Jingshan Pan$^3$, Wen Gao$^1$ \\
    \normalsize $^1$National Research Engineering Center of Visual Technology, School of Computer Science, Peking University, Beijing, China \\
    \normalsize $^2$University of Chinese Academy of Sciences, Beijing, China \\
    \normalsize $^3$National Supercomputer Center in Jinan, Jinan, China \\
    \footnotesize\texttt{
        qizhang@stu.pku.edu.cn, \{sswang,swma,wgao\}@pku.edu.cn, xfzhang@ucas.ac.cn, panjsh@sdas.org
    }
}

\maketitle

\begin{abstract}
   Nowadays, high-quality images are pursued by both humans for better viewing experience and by machines for more accurate visual analysis. However, images are usually compressed before being consumed, decreasing their quality. It is meaningful to predict the perceptual quality of compressed images for both humans and machines, which guides the optimization for compression. In this paper, we propose a unified approach to address this. Specifically, we create a deep learning-based model to predict Satisfied User Ratio (SUR) and Satisfied Machine Ratio (SMR) of compressed images simultaneously. We first pre-train a feature extractor network on a large-scale SMR-annotated dataset with human perception-related quality labels generated by diverse image quality models, which simulates the acquisition of SUR labels. Then, we propose an MLP-Mixer-based network to predict SUR and SMR by leveraging and fusing the extracted multi-layer features. We introduce a Difference Feature Residual Learning (DFRL) module to learn more discriminative difference features. We further use a Multi-Head Attention Aggregation and Pooling (MHAAP) layer to aggregate difference features and reduce their redundancy. Experimental results indicate that the proposed model significantly outperforms state-of-the-art SUR and SMR prediction methods. Moreover, our joint learning scheme of human and machine perceptual quality prediction tasks is effective at improving the performance of both.
\end{abstract}


\section{Introduction}

A large amount of images are consumed by not only humans but also machines every day. Image quality is an important attribute that significantly affects the user experience and machine analysis performance, and thereby higher image quality is preferred by both. However, images are usually compressed before spread, and the compression process inevitably degrades the image quality. It is essential to assess the compressed image quality so that more efficient compression methods can be developed according to the assessment result.

In the past decades, many image quality assessment (IQA) models have been proposed to predict the image quality perceived by humans \cite{zhai2020perceptual}. Traditional metrics, like PSNR and SSIM, are designed based on the pixel-wise difference between the reference and distorted images. These metrics are applied widely, but are not well correlated with human perception. Recently, many deep learning-based IQA metrics \cite{zhang2018unreasonable,ding2020image,chen2024topiq} have been developed. Most of them utilize the difference of extracted deep features to predict a perceptual quality score. Since these features are learned to contain more task-related information, deep IQA metrics have outperformed traditional ones, and are becoming more broadly-used nowadays.

However, existing IQA metrics seldom model the characteristics of the human vision system (HVS), making them still inconsistent and unreliable, especially when the compressed image quality varied subtlely \cite{zhang2021fine}. Specifically, HVS has inherent flaws that can only perceive quality difference beyond a threshold, which is called Just Noticeable Difference (JND) \cite{wu2019survey}. By collecting many JND samples from different humans, the Satisfied User Ratio (SUR) of a compressed image can be calculated \cite{wang2017videoset}, which is the ratio of subjects who cannot perceive the quality difference between the original image and compressed one. Compared with other aforementioned IQA metrics, SUR directly reflects the HVS characteristics, and is more consistent with general human perception. Several methods have been developed to predict the SUR of compressed images \cite{fan2019net,lin2020featnet}. However, the prediction error can be further decreased by improving the network architecture.

Different from humans, machines perceive images by understanding their contents. Some recent works reveal that machine vision systems (MVS) also have JND characteristics \cite{zhang2021just}, and the machine diversity is also addressed in \cite{zhang2024perceptual}, where the concept of Satisfied Machine Ratio (SMR) is introduced. Similar to SUR, SMR is defined as the ratio of machines that cannot perceive the quality difference between the original image and compressed one. Therefore, it is reasonable to evaluate the image quality for machines by predicting its SMR, which has not been well studied yet.

In many applications, images are consumed by large populations of both humans and machines. It is meaningful to predict SUR and SMR simultaneously, which can guide the development of better compression methods to satisfiy both kinds of consumers. Since SUR and SMR are both related to the perceptual quality of images, leveraging such relationship and designing a unified model to predict them can be more effective than predicting them separately. In this paper, we propose a learning-based model to achieve this. Our contributions can be summarized as follows:


\begin{itemize}
    \item We propose a unified model to predict SUR and SMR of compressed images simultaneously. The joint learning of HVS- and MVS-perception related features implicitly leads to more effective feature extraction for both tasks.
    \item We introduce a pre-train scheme that does not rely on large-scale SUR-annotated data, which is hard to obtain. Instead, we use diverse IQA models to generate normalized quality labels for compressed images, which aligns with the annotation process of SUR labels.
    \item We introduce a Difference Feature Residual Learning (DFRL) module to learn more discriminative difference features. We then use a Multi-Head Attention Aggregation and Pooling (MHAAP) layer to aggregate the multi-layer difference features and pool them into a smaller representation. Moreover, we propose to use an MLP-Mixer network to fuse the spatial and channel information of the aggregated representation for SUR and SMR prediction.
    \item Experimental results show that the proposed model outperforms state-of-the-art SUR and SMR prediction models. And the joint learning scheme of SUR and SMR prediction tasks can improve the performance of both.
\end{itemize}

\section{Revisit SUR and SMR}

We first give a unified formulation of SUR and SMR. Let $I_0$ be the original image, and $I_{q_1}, I_{q_2}, \ldots, I_{q_K}$ be its compressed versions. The image quality is controlled by compression parameters $q_k$, which degrades as $k$ increases. For any human or machine subject $X$, we use the following function to determine where it is satisfied with the image quality of $I_{q_k}$:

\begin{equation}
    S(X; I_{q_k}) = 
  \begin{cases}
    1, & \text{if}\ P(X, I_0) = P(X, I_{q_k}) \\
    0, & \text{otherwise}
  \end{cases},
\end{equation}

\noindent where $P(\cdot)$ is the perception result of $X$ on the image. For humans, a subjective test can be conducted to check and record whether participants can perceive the quality difference between $I_0$ and $I_{q_k}$ or not \cite{wang2017videoset}. For machines, we can run them on $I_0$ and $I_{q_k}$ and know if the two results are the same or similar enough \cite{zhang2024perceptual}. Then, by collecting $S(X; I_{q_k})$ from a large population of $X$, SUR or SMR (jointly denoted as $R$) can be calculated as

\begin{equation}
    R(I_{q_k}) = \frac{| \{X \in \mathbb{X} \mid S(X; I_{q_k}) = 1\} |}{|\mathbb{X}|},
\end{equation}

\noindent where $\mathbb{X}$ is the set of subjects and $|\cdot|$ counts its elements. Therefore, SUR and SMR of $I_{q_k}$ are the ratios of subjects who are satisfied with its quality. Because of such acquisition process, SUR and SMR reflect the general perceptual characteristics of humans and machines instead of individuals, removing the subject bias and making them more consistent and reliable. This advantage is particularly significant since the ultimate receivers of images are often large and diverse populations of humans or machines. SUR and SMR are also more direct and promising quality indicators than other IQA metrics, because in practice images are compressed to satisfy some percentage of humans or machines.

\begin{figure}
    \centering
    \subfloat{
        \includegraphics[width=0.48\textwidth]{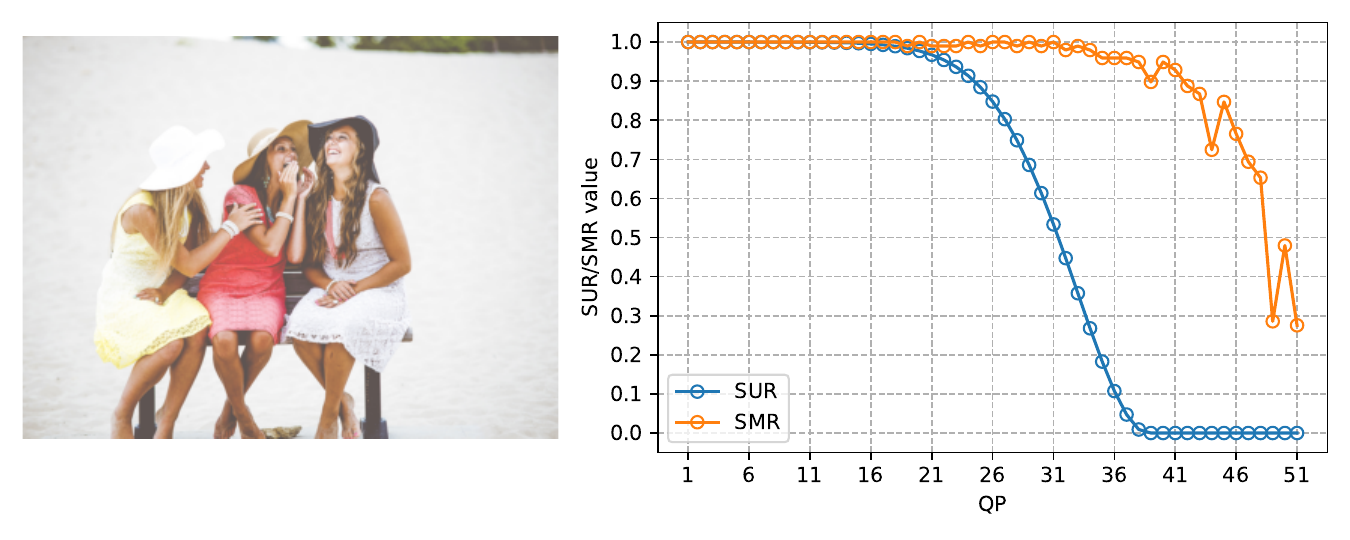}
    }
    \quad
    \subfloat{
        \includegraphics[width=0.48\textwidth]{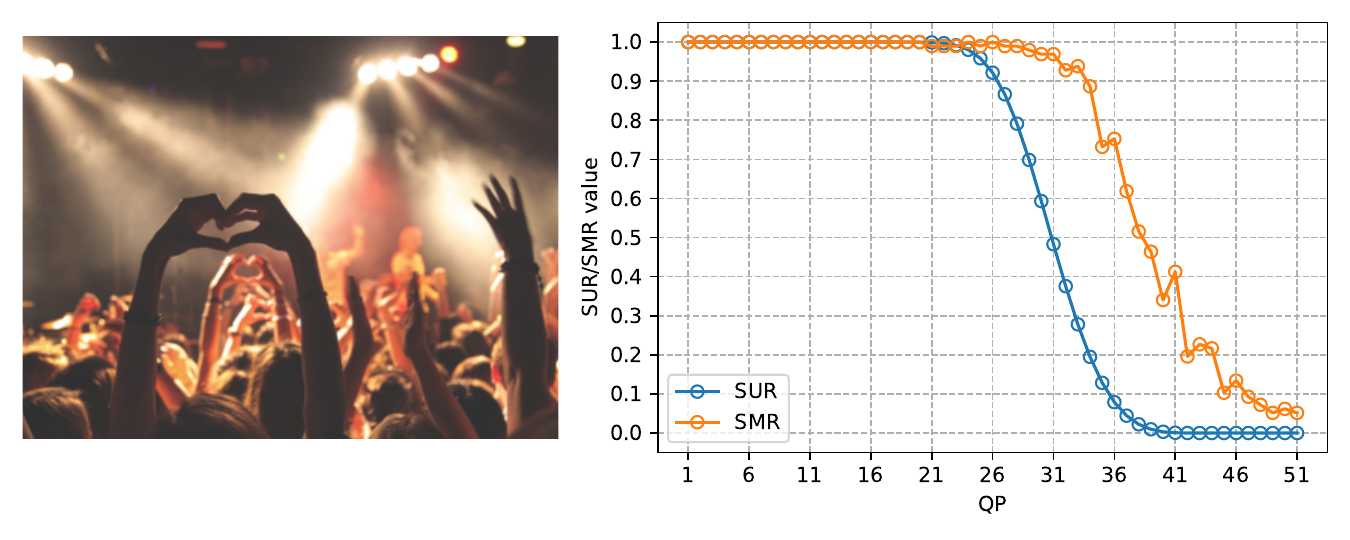}
    }
    \caption{SUR and SMR distributions of two sample images in the KonJND-1k dataset.}
    \label{fig:qp-sursmr-samples}
\end{figure}

In Fig. \ref{fig:qp-sursmr-samples}, we show SUR and SMR distributions under different compression ratios (controlled by quantization parameter) of two images in the KonJND-1k dataset \cite{lin2022large}. For different images, it is obvious that both distributions are distinct. For the same image, the SUR distribution exhibits some more favorable properties than SMR: i) it is always monotonically non-increasing, and ii) it always starts from 1.0 (no one can perceive the quality difference) and ends with 0.0 (everyone can perceive the quality difference). On the contrary, SMR curves usually have huge variations even between neighboring QPs (many machines are not robust to certain distortion levels \cite{zhang2024perceptual}) and are not always falling to 0.0 at high QPs (some machines can be extremely robust to large distortions).

SUR and SMR prediction tasks have several challenges:

\begin{itemize}
    \item \textbf{Lack of large-scale dataset.} The scale of SUR datasets is significantly limited due to the high cost of subjective tests. Currently, the largest SUR dataset KonJND-1k \cite{lin2022large} contains 1,008 original images and 76k compressed images, which is much smaller than popular machine vision task datasets like MS-COCO. Although the compression process can produce more variants of an original image, the labeled data is still insufficient. On the other hand, the scale of SMR dataset is larger: the MS-COCO based SMR dataset (COCO-SMR) proposed in \cite{zhang2024perceptual} contains more than 123k original images and 4 million compressed ones with annotations. However, this dataset does not provide SUR labels.
    \item \textbf{Modern and effective network architecture.} For the SUR prediction task, the state-of-the-art (SOTA) method \cite{lin2020featnet} uses an Inception-V3 network \cite{szegedy2016rethinking} to extract multi-layer features from original and compressed images, and then predicts SMR by an MLP regression network. For SMR prediction, the SOTA method \cite{zhang2024perceptual} adopts an EfficientNet \cite{tan2019efficientnet} as the backbone, and only leverages the highest level of features. These networks are not the most capable ones nowadays, so the prediction errors have the potential to be further decreased with the latest architectures like Transformer \cite{dosovitskiy2020image} or MLP-Mixer \cite{tolstikhin2021mlp}.
    \item \textbf{Joint feature learning of general HVS and MVS characteristics.} Since SUR and SMR are both related to the perceptual quality of images, it is reasonable to design a network to predict them at the same time. Intuitively, HVS pursues the fidelity of low-level textures and details after compression, while MVS focuses more on the preservation of high-level semantics. However, as revealed in \cite{zhang2018unreasonable}, the features learned towards MVS-targetted tasks are also effective at assessing image perceptual quality for HVS. Similarly, the difference of low-level information between original and compressed images can also be beneficial for SMR prediction.
    \item \textbf{Fine granularity and high precision.} SUR and SMR datasets provide annotations at a fine granularity, which means that the quality difference between neighboring compressed images is subtle. Therefore, SUR and SMR prediction tasks are more difficult than traditional IQA tasks, which usually predict image quality with larger distortion level gaps and thus is easier \cite{zhang2021fine}. Furthermore, good IQA metrics only need to be well correlated with ground truth labels like mean opinion scores, while SUR and SMR prediction models need to predict the exact SUR and SMR values, which requires higher precision.
\end{itemize}

In this paper, we address these challenges by proposing a unified model to predict SUR and SMR of compressed images simultaneously. This model is based on several capable network modules, and pre-trained on a large-scale dataset with SUR- and SMR-related annotations.

\section{Method}


\begin{figure*}[htbp]
    \centering
    \includegraphics[width=0.96\textwidth]{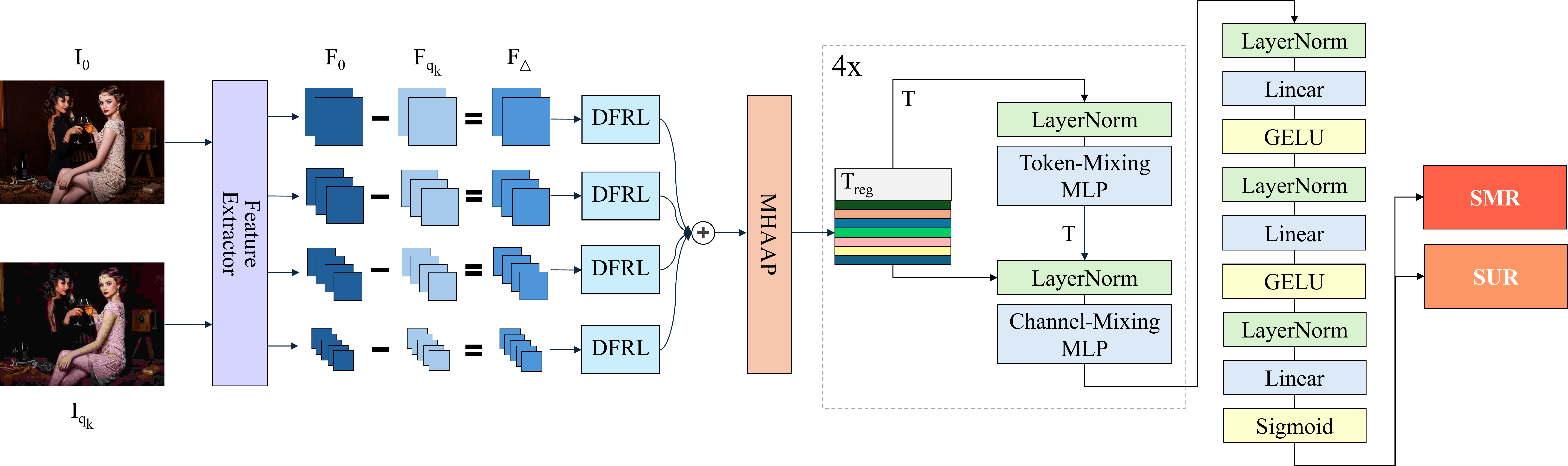}
    \caption{The proposed SUR and SMR prediction network. We first extract multi-layer original features $F_0$ and compressed features $F_{q_k}$ by a CAFormer, then the initial difference features $F_\Delta$ are input to the Difference Feature Residual Learning (DFRL) module to generate more discriminative difference features $F^{*}_\Delta$. Subsequently, $F^{*}_\Delta$ are aggregated and pooled by the Multi-Head Attention Aggregation and Pooling (MHAAP) layer to obtain $F^{*,attn}_\Delta$. Finally, $F^{*,attn}_\Delta$ is concatenated with a regression token $T_\text{reg}$,fused by 4 MLP-Mixer layers, and fed into another 3-layer MLP to predict SUR and SMR.}
    \label{fig:proposed network}
\end{figure*}

\subsection{Large-scale joint pre-training via proxy task}

In \cite{zhang2024perceptual}, a large SMR dataset is established, which contains enough annotated images to pre-train an SMR prediction model. However, it is impractical to annotate SUR for each of the 123k images in it. To take more advantage of this database and facilitate joint learning of SUR and SMR prediction, we need to generate SUR-related labels for these images and design a proxy task to learn HVS perceptual characteristics. Since the SUR prediction task is similar to IQA, and the current best-performing full-reference (FR) IQA models can effectively evaluate image quality, we can use an FR-IQA model to output quality scores as SUR-related labels. This approach has already been adopted in \cite{lin2020featnet}.

In this paper, we move a step forward to generate more consistent and reliable SUR-related labels. Since the SUR is calculated by aggregating perception results from different human subjects, we simulate this process by choosing a number of IQA models, collecting their predicted quality scores, and generating an averaged and normalized quality score at last. Specifically, let $I_0$ and $I_{q_k}$ be the original image and its compressed version controlled by parameter $q_k$, we select $N$ IQA models $M_1, M_2, \ldots, M_N$ for the annotation. The SUR-related label $\hat{SUR}$ of image $I_{q_k}$ is calculated as

\begin{equation}
    \hat{SUR}(I_{q_k}) = \frac{1}{N} \sum_{i=1}^{N} \frac{M_i(I_{q_k}) - \min(M_i(\mathbb{I}))}{\max(M_i(\mathbb{I})) - \min(M_i(\mathbb{I}))},
\end{equation}

\noindent where $\mathbb{I}=I_{q_1}, I_{q_2}, \ldots, I_{q_K}$, and $\min(\cdot)$ and $\max(\cdot)$ return the minimum and maximum value, respectively. This equation assumes that the quality score is the higher, the better. If an IQA model disobeys this rule, we simply reverse its output by $1 - M_i(I_{q_k})$. We carefully select 14 representative FR-IQA models for labeling, which are SSIM \cite{wang2004image}, MS-SSIM \cite{wang2003multiscale}, VIF \cite{sheikh2006image}, CW-SSIM \cite{sampat2009complex}, FSIM \cite{zhang2011fsim}, LPIPS-Alex and LPIPS-VGG \cite{zhang2018unreasonable}, PieAPP \cite{prashnani2018pieapp}, DISTS \cite{ding2020image}, CKDN \cite{zheng2021learning}, ST-LPIPS-Alex and ST-LPIPS-VGG \cite{ghildyal2022shift}, and TOPIQ and TOPIQ-Pipal \cite{chen2024topiq}. Both traditional and deep learning-based models are included to increase the diversity. We find it quite common that $M_i(I_{q_k}) \leq M_i(I_{q_{k+\epsilon}})$ for any selected $M_i$, \textit{i.e.} the compression becomes heavier while the quality score increases, suggesting the inconsistency of existing IQA models. By introducing more models and aggregating their predictions, this issue has been alleviated. The distribution of the averaged normalized quality scores are also closer to SUR.

Subsequently, a feature extractor network $G$ is trained on this dataset, which receives original and compressed images as inputs, and predicts $\hat{SUR}$ and SMR simultaneously. Hence, the SUR prediction task is proxied by $\hat{SUR}$ prediction, which learns appropriate HVS perceptual characteristics-related features. The training objective is to minimize the L1 loss between predicted and ground-truth $\hat{SUR}$ and SMR values:

\begin{equation}
    \min \mathcal{L} = \alpha \cdot |G_{\hat{SUR}} - \hat{SUR}| + \beta \cdot |G_{SMR} - SMR|,
\end{equation}

\noindent where $\alpha=\beta=0.5$ are weights of two losses.

\subsection{SUR and SMR prediction network}

Previous SUR and SMR prediction models are mostly using convolutional neural networks (CNNs) as backbones \cite{lin2020featnet,zhang2024perceptual}. However, it is demonstrated that capturing HVS perceptual characteristics requires the network to focus on both low-level details and high-level semantics \cite{xu2024boosting}, which are better modeled by CNNs and Transformers, respectively. On the other hand, the MVS perceptual characteristics are related to the architectures of the evaluated machines. Typically, both CNNs and Transformers should be considered because of their utility in different scenarios \cite{zhang2024perceptual}. Therefore, in this work, we propose an SUR and SMR prediction network that leverages both convolution operation and attention mechanism.

Specifically, in the pre-training stage, we use a CAFormer network \cite{yu2023metaformer} as the feature extractor, which follows a four-stage scheme. A convolution block is used in the first two stages as a mixer of feature sequences (or called tokens), while a self-attention block is used in the last two stages. Therefore, both local texture and global semantic information can be modeled. Finally, the extracted original and compressed features are used to calculate the feature difference, which are then concatenated as the input of a 3-layer MLP to predict $\hat{SUR}$ and SMR.

The pre-trained network is rather simple for feature extraction, while in the fine-tuning stage, we use a more complicated and capable network to fully exploit the extracted features. It is common to combine original multi-layer features $F_0^l$ (where $l$ is the layer index), compressed features $F_{q_k}^{l}$, and their difference $F_\Delta^{l}$ to predict SUR \cite{lin2020featnet} or SMR \cite{zhang2024perceptual}. However, these multi-layer and multi-quality representations are often noisy and in a large data volume, lowering the model's efficiency and performance. To address this, we propose to only leverage the difference features $F_\Delta$ for our task with proper feature selection and fusion.

Commonly, difference features are calculated as $F_\Delta^{l} = F_0^{l} - F_{q_k}^{l}$ \cite{lin2020featnet,zhang2024perceptual,chen2024topiq}. This per-pixel subtraction operation is straightforward but limited in representation capability, particularly when we should capture both HVS and MVS perceptual characteristics. We introduce a \textbf{Difference Feature Residual Learning (DFRL)} module to increase such capability, which is a 3-layer convolutional network that can be formulated as

\begin{equation}
    F_\Delta^{*,l} = F_\Delta^{l} + \text{Conv}^l_3(\sigma(\text{Conv}^l_2(\sigma(\text{Conv}^l_1(F_\Delta^{l}))))),
\end{equation}

\noindent where $\sigma(\cdot)$ is the GELU function and $\text{Conv}^l_i$ is the $i$-th convolutional layer. By introducing non-linear transformations, the DFRL module learns more discriminative difference features for both SUR and SMR prediction. It can also reduce the noise in low-level features via learnable filtering and purifying.

Then, we design a \textbf{Multi-Head Attention Aggregation and Pooling (MHAAP)} layer to aggregate the multi-layer difference features and pool them into a much smaller representation via adaptive attention. Firstly, the learned multi-layer difference features are interpolated to the same spatial size $H \times W$ and concatenated by channel-wise, which forms a latent feature $F^{*}_\Delta \in \mathbb{R}^{C \times H \times W}$, where $C$ is the sum of channels of different layers. Secondly, $F^{*}_\Delta$ is reshaped and normalized as a sequence of $N = H \times W$ spatial tokens. To better aggregate global information from these tokens and reduce noise or redundancy, we introduce a learnable query vector $Q \in \mathbb{R}^{Y \times C}$, where $Y \ll N$. Thirdly, we project $F^{*}_\Delta$ into key ($K$) and value $V$ matrices as $K = F^{*}_\Delta W_K$ and $V = F^{*}_\Delta W_V$. After that, let $h$ be the number of heads, the multi-head attention mechanism is applied to $Q, K, V$ to calculate the attention weights:

\begin{equation}
    \text{Attention}(Q, K, V) = \text{Softmax}(\frac{QK^T}{\sqrt{C/h}})V.
\end{equation}

\noindent Finally, the attention-weighted difference features are pooled by a simple linear layer to obtain the aggregated representation $F^{*,attn}_\Delta \in \mathbb{R}^{Y \times Z}$, where $Z < C$. This representation contains both low-level and high-level information of the difference features, which is beneficial for simultaneously modeling the HVS and MVS perceptual characteristics on distinguishing the quality difference between original and compressed images.

To further fuse the spatial and channel information of $F^{*,attn}_\Delta$, we propose to use an MLP-Mixer network \cite{tolstikhin2021mlp}. Specifically, we introduce a learnable regression token $T_{reg} \in \mathbb{R}^{1 \times Z}$ as a global context assembler, which is concatenated with $F^{*,attn}_\Delta$ to form the input of the MLP-Mixer as $S$. The mixing process can be formulated as

\begin{equation}
    \begin{split}
        U &= S + (W_2 \sigma(W_1 \text{LayerNorm}(S^T)))^T, \\
        V &= U + (W_4 \sigma(W_3 \text{LayerNorm}(U))),
    \end{split}
\end{equation}

\noindent where $W_*$ are learnable weights, $\sigma(\cdot)$ is the GELU function, and $*^T$ denotes the transpose operation. Finally, the mixing result stored in $T_{reg}$ is fed into another 3-layer MLP to predict SUR and SMR. The whole network architecture is shown in Fig. \ref{fig:proposed network}, and the training objective is similar to Eq. (4) while using the ground truth SUR labels.

\section{Experiments}

\subsection{Experimental setup}

We use the COCO-SMR dataset for the pre-training. After that, we use MCL-JCI \cite{jin2016statistical}, VVC-JND \cite{shen2020jnd}, and KonJND-1k \cite{lin2022large} to fine-tune our model and evaluate the performance, which contains 50, 1008, and 202 original images, respectively. Note that the compression types are different among these datasets, which are JPEG for \cite{jin2016statistical}, VVC for \cite{shen2020jnd}, and JPEG or BPG for \cite{lin2022large}. The compression levels are correspondingly different: 100 for \cite{jin2016statistical}, 39 for \cite{shen2020jnd}, and 100 or 51 for \cite{lin2022large}. Each dataset will be randomly split into training, validation, and test sets with a ratio of 7:1:2. Since they all lack SMR annotations, we follow the same procedure in \cite{zhang2024perceptual} to annotate SMR scores for the object detection task on them. We also mix these datasets into a larger one, called as SUR-SMR dataset, for more comprehensive evaluation.

We compare our methods with state-of-the-art SUR and SMR prediction models, \textit{i.e.} SUR-FeatNet \cite{lin2020featnet} and SMR-Net \cite{zhang2024perceptual}. For SUR-FeatNet, we use their official weights for extracting features and fine-tune the SUR regression network on the evaluated datasets. For SMR-Net, the whole network is initialized using official weights trained on the COCO-SMR dataset and fine-tuned on the evaluated datasets. The SUR and SMR prediction performance is evaluated by the mean absolute error (MAE) between predicted and ground-truth values, \textit{i.e.} $|\Delta \text{SUR}|$ and $|\Delta \text{SMR}|$. Note that we directly predict SUR and SMR for the whole image instead of predicting them at patch-level and then aggregating the results like in \cite{lin2020featnet}, which is more reasonable yet challenging. The input images of all models are in the resolution of $224\times224$, which is also different from that in \cite{zhang2024perceptual} ($512\times512$). During training and fine-tuning, we adopt mixed precision for acceleration and use the Adam optimizer with a learning rate of $10^{-4}$. The batch size is 72 on three RTX 4090 GPUs. We adopt random horizontal and vertical flipping as the only data augmentation technique during training. Besides, we set $Y=7 \times 7, h=8, Z=512$.


\subsection{Main results}

\bgroup
\def\arraystretch{1.36}
\begin{table}[htbp]
    \centering
    \caption{Main results.}
    \label{tab:main results}
    \resizebox{0.48\textwidth}{!}{
    \begin{tabular}{c|cc|cc}
    \Xhline{4\arrayrulewidth}
    \textbf{} & \multicolumn{2}{c|}{\textbf{$|\Delta \text{SUR}|$}} & \multicolumn{2}{c}{\textbf{$|\Delta \text{SMR}|$}} \\ \hline
    \textbf{Dataset} & \multicolumn{1}{c}{\textbf{Ours}} & \textbf{\cite{lin2020featnet}} & \multicolumn{1}{c}{\textbf{Ours}} & \textbf{\cite{zhang2024perceptual}} \\
    \Xhline{4\arrayrulewidth}
    \textbf{KonJND-BPG} & \multicolumn{1}{c}{\textbf{0.0339}} & 0.0704 & \multicolumn{1}{c}{\textbf{0.0352}} & 0.0437 \\ \hline
    \textbf{KonJND-JPEG} & \multicolumn{1}{c}{\textbf{0.0655}} & 0.1062 & \multicolumn{1}{c}{\textbf{0.0309}} & 0.0378 \\ \hline
    \textbf{VVC-JND} & \multicolumn{1}{c}{\textbf{0.0740}} & 0.1190 & \multicolumn{1}{c}{\textbf{0.0530}} & 0.0765 \\ \hline
    \textbf{MCL-JCI} & \multicolumn{1}{c}{\textbf{0.0529}} & 0.1113 & \multicolumn{1}{c}{\textbf{0.0319}} & 0.0648 \\ \hline
    \textbf{SUR-SMR} & \multicolumn{1}{c}{\textbf{0.0570}} & 0.1096 & \multicolumn{1}{c}{\textbf{0.0331}} & 0.0392 \\
    \Xhline{4\arrayrulewidth}
    \end{tabular}
    }
\end{table}
\egroup

The main experimental results are shown in TABLE \ref{tab:main results}. It is clear that our method significantly outperforms SOTA methods in predicting both SUR and SMR on all datasets. The prediction errors can be decreased by at most $0.0584$ and $0.0329$, respectively, both on the MCL-JCI dataset. We find that SMR prediction performance gap between our method and \cite{zhang2024perceptual} decreases consistently as the scale of dataset increases. On the mixed SUR-SMR dataset, the prediction errors are decreased by $0.0526$ and $0.0061$ compared to SOTA methods. These results demonstrate the effectiveness of our method in predicting SUR and SMR of compressed images.

\bgroup
\def\arraystretch{1.36}
\begin{table}[htbp]
    \centering
    \caption{Model complexity comparison.}
    \label{tab:model complexity}
    \resizebox{0.36\textwidth}{!}{
    \begin{tabular}{c|ccc}
    \Xhline{4\arrayrulewidth}
    & \textbf{Ours} & \textbf{\cite{lin2020featnet}} & \textbf{\cite{zhang2024perceptual}} \\ 
    \Xhline{3\arrayrulewidth}
    \textbf{Model params (M)} & 47.00 & 49.02 & 31.46 \\ \hline
    \textbf{GFLOPS} & 22.56 & 3.19 & 5.74 \\ \hline
    \textbf{Infer time (ms)} & 36.96 & 35.03 & 20.73 \\ 
    \Xhline{4\arrayrulewidth}
    \end{tabular}
    }
\end{table}
\egroup

We also compare the model complexity of our method with SOTA methods in TABLE \ref{tab:model complexity}. Because we introduce several more advanced network modules like multi-head attention and MLP-Mixer, our model exhibits a higher computational cost with 22.56 GFLOPS than \cite{lin2020featnet} and \cite{zhang2024perceptual}. However, the model size and inference time are still comparable and acceptable, which are 47.00M parameters and 36.96ms, respectively. More importantly, our approach of using a unified model to predict SUR and SMR simultaneously can also reduce the model complexity compared to using separate models for each task.

\subsection{Ablation study}

\bgroup
\def\arraystretch{1.36}
\begin{table}[htbp]
    \centering
    \caption{SUR and SMR prediction performance under different pre-training and fine-tuning strategies.}
    \label{tab:ablation study 1}
    \resizebox{0.36\textwidth}{!}{
    \begin{tabular}{c|cc}
    \Xhline{4\arrayrulewidth}
    & \textbf{$|\Delta \text{SUR}|$} & \textbf{$|\Delta \text{SMR}|$} \\ 
    \Xhline{3\arrayrulewidth}
    \textbf{Ours$_\text{SUR+SMR}$} & \textbf{0.0570} & \textbf{0.0331} \\ \hline
    \textbf{Ours$_\text{Scratch}$} & 0.0702 & 0.0469 \\ \hline
    \textbf{Ours$_\text{SUR}$} & 0.0584 & - \\ \hline
    \textbf{Ours$_\text{SMR}$} & - & 0.0346 \\ \hline
    \textbf{Ours$_\text{LPIPS+SMR}$} & 0.0581 & 0.0351 \\ \hline
    \textbf{Ours$_\text{DISTS+SMR}$} & 0.0581 & 0.0347 \\
    \Xhline{4\arrayrulewidth}
    \end{tabular}
    }
\end{table}
\egroup

We want to verify the effectiveness of the joint learning scheme proposed in this work. We first keep the network architecture unchanged and direct train it to predict SUR and SMR on the SUR-SMR dataset without any pre-training. The result is shown in TABLE \ref{tab:ablation study 1}, annotated by ``Ours$_\text{Scratch}$''. Obviously, joint pre-training can significantly decrease the prediction errors for both tasks by $0.0132$ and $0.0138$, respectively. It is worth mentioning that images in the COCO-SMR dataset are compressed with HEVC, and thereby the distortion is different from that of the SUR-SMR dataset, while the pre-trained network can still be fine-tuned effectively on the latter dataset. Then, we pre-train and fine-tune our networks separately for SUR and SMR prediction tasks. In TABLE \ref{tab:ablation study 1}, ``Ours$_\text{SUR}$'' means pre-training with only $\hat{SUR}$ labels and fine-tuning with only SUR labels, and ``Ours$_\text{SMR}$'' means pre-training and fine-tuning with only SMR labels. The results show that the joint pre-training scheme is effective for both tasks. Specifically, SUR and SMR prediction errors are decreased by $0.0014$ and $0.0015$, respectively.

The $\hat{SUR}$ annotation process is designed to remove the inconsistency and perception bias from a single IQA model. To verify this, we compare our approach with two variants, \textit{i.e.} ``Ours$_\text{LPIPS+SMR}$'' and ``Ours$_\text{DISTS+SMR}$'', which use LPIPS and DISTS as the only IQA model to annotate SUR-related quality score for joint pre-training, respectively. According to the comparison results in TABLE \ref{tab:ablation study 1}, these two variants bring performance drops for both SUR and SMR prediction, suggesting the superiority of our approach.

\bgroup
\def\arraystretch{1.36}
\begin{table}[htbp]
    \centering
    \caption{SUR and SMR prediction performance under different model architectures.}
    \label{tab:ablation study 2}
    \resizebox{0.36\textwidth}{!}{
    \begin{tabular}{c|cc}
    \Xhline{4\arrayrulewidth}
    & \textbf{$|\Delta \text{SUR}|$} & \textbf{$|\Delta \text{SMR}|$} \\ 
    \Xhline{3\arrayrulewidth}
    \textbf{Ours} & \textbf{0.0570} & \textbf{0.0331} \\ \hline
    \textbf{Ours$_\text{w/o DFRL}$} & 0.0627 & 0.0343 \\ \hline
    \textbf{Ours$_\text{w/o MHAAP}$} & 0.0585 & 0.0344 \\ \hline
    \textbf{Ours$_\text{Transformer}$} & 0.0583 & 0.0338 \\ \hline
    \textbf{Ours$_\text{MLP}$} & 0.0584 & 0.0338 \\ \hline
    \textbf{Ours$_\text{all features}$} & 0.0584 & 0.0331 \\ 
    \Xhline{4\arrayrulewidth}
    \end{tabular}
    }
\end{table}
\egroup

Finally, we evaluate the performance of different network architectures. In TABLE \ref{tab:ablation study 2}, Ours$_\text{w/o DFRL}$ means replacing the DFRL module with $F_\Delta^{l} = F_0^{l} - F_{q_k}^{l}$. Ours$_\text{w/o MHAAP}$ means replacing the MHAAP layer with a $1\times1$ convolutional layer and a global average pooling layer for channel- and spatial-wise pooling. According to the results, both modules contribute to the prediction error decreasement. Furthermore, we replace 4 MLP-Mixer layers with 4 Transformer encoder layers (8 heads) or merely 1 MLP layer ($Y \times Z$ neurons). The results show that the MLP-Mixer structure is more effective in fusing aggregated and pooled difference features. We also compare the performance of using all features ($F_0, F_{q_k}$, and $F^{*}_\Delta$) for prediction with using only $F^{*}_\Delta$. The results denoted by ``Ours$_\text{all features}$'' in TABLE \ref{tab:ablation study 2} show that this alternative increases the SUR prediction error, and it also adds additional computational cost, which is not beneficial at all.

\section{Conclusion}

In this paper, we propose a unified SUR and SMR prediction model for compressed images. To address the lack of large-scale SUR-annotated data, we use a series of IQA models to generate normalized quality scores as proxy lables for compressed images in a MS-COCO based SMR dataset. Then, we pre-train a CAFormer-based feature extractor network to learn HVS and MVS perceptual characteristics. Subsequently, we design an MLP-Mixer-based model to predict SUR and SMR by leveraging and fusing multi-layer difference features, which are learned by a Difference Feature Residual Learning (DFRL) module and aggregated by a Multi-Head Attention Aggregation and Pooling (MHAAP) layer. Experimental results show that our model outperforms state-of-the-art SUR and SMR prediction models, and the joint learning scheme improves the prediction performance of both.

\bibliographystyle{IEEEbib}
\bibliography{icme2025references}

\end{document}